\documentclass[preprint,12pt]{elsarticle}

\usepackage{amssymb}
\usepackage{graphicx}%
\usepackage{multirow}%
\usepackage{amsmath,amssymb,amsfonts}%
\usepackage{amsthm}%
\usepackage{mathrsfs}%
\usepackage[title]{appendix}%
\usepackage{xcolor}%
\usepackage{textcomp}%
\usepackage{manyfoot}%
\usepackage{booktabs}%
\usepackage{algorithm}%
\usepackage{algorithmicx}%
\usepackage{algpseudocode}%
\usepackage{listings}%
\usepackage{tabularx}
\journal{Nuclear Physics B}

\begin{document}

\begin{frontmatter}



\title{Trustworthy Privacy-Preserving Multimodal Federated Learning for Personalised Breast Cancer Prediction}


\author[ntu]{Ruth Amey} %% Author name
% \ead{ruth.amey@icloud.com}
\author[ntu]{Muhammad Arifur Rahman}
\ead{arif.rahman@ntu.ac.uk}
\ead[url]{https://www.ntu.ac.uk/staff-profiles/science-technology/arif, arif.rahman@ntu.ac.uk}
\author[ntu]{Taha Osman}
\author[ntu]{Nicholas Shopland}
\author[ntu]{Andy Burton}
\author[kfupm]{Mufti Mahmud}
\author[ntu]{David J. Brown}

%\author[1,2]{Han Thane\corref{cor1}%\fnref{fn1}}
%\ead{han@different.edu}
%\author[3]{T. Rafeeq\fnref{fn1,fn3}}
%\fntext[fn1]{This is the first author footnote.}
%\fntext[fn2]{Another author footnote, this is a very long footnote and
%it should be a really long footnote. But this footnote is not yet
%sufficiently long enough to make two lines of footnote text.}
\affiliation[ntu]{
 organization={Department of Computer Science},
  addressline={Nottingham Trent University},
  city={Nottingham},
  country={UK}
}

\affiliation[kfupm]{
 organization={Department of Information and Computer Science\\ SDAIA-KFUPM Joint Research Center for AI \\ IRC for Bio Systems and Machines},
  addressline={King Fahd University of Petroleum \& Minerals},
  city={},
  country={Saudi Arabia}
}

% \author{} %% Author name

% \affiliation{organization={},%Department and Organization

%% Abstract
\begin{abstract}
Federated learning has emerged as a potential solution to privacy concerns associated with using sensitive health data for training predictive models, particularly in personalised cancer care. This research investigates whether federated learning can support the development of robust models for predicting tumour progression in breast cancer patients while addressing four critical deployment pillars: transparency, scalability, security, and fairness. This study evaluates a federated learning framework using multimodal data, including clinical information, tumour characteristics, biomarker data, and patient demographics, alongside medical imaging data such as MRI scans, to model changes in tumour characteristics over time. The performance of the federated approach was compared with that of a centralised model trained on aggregated data. The report then further examines strategies to enhance secure model updates, maintain performance across patient subgroups, and support scalability across institutions. The findings assess whether federated learning can achieve predictive performance comparable to centralised learning while preserving data locality. These results contribute to understanding the feasibility of privacy-preserving, multimodal predictive modelling and support future applications such as digital twins to assist clinicians and patients in personalised treatment planning.
\end{abstract}

\begin{graphicalabstract}
\includegraphics[width=\linewidth]{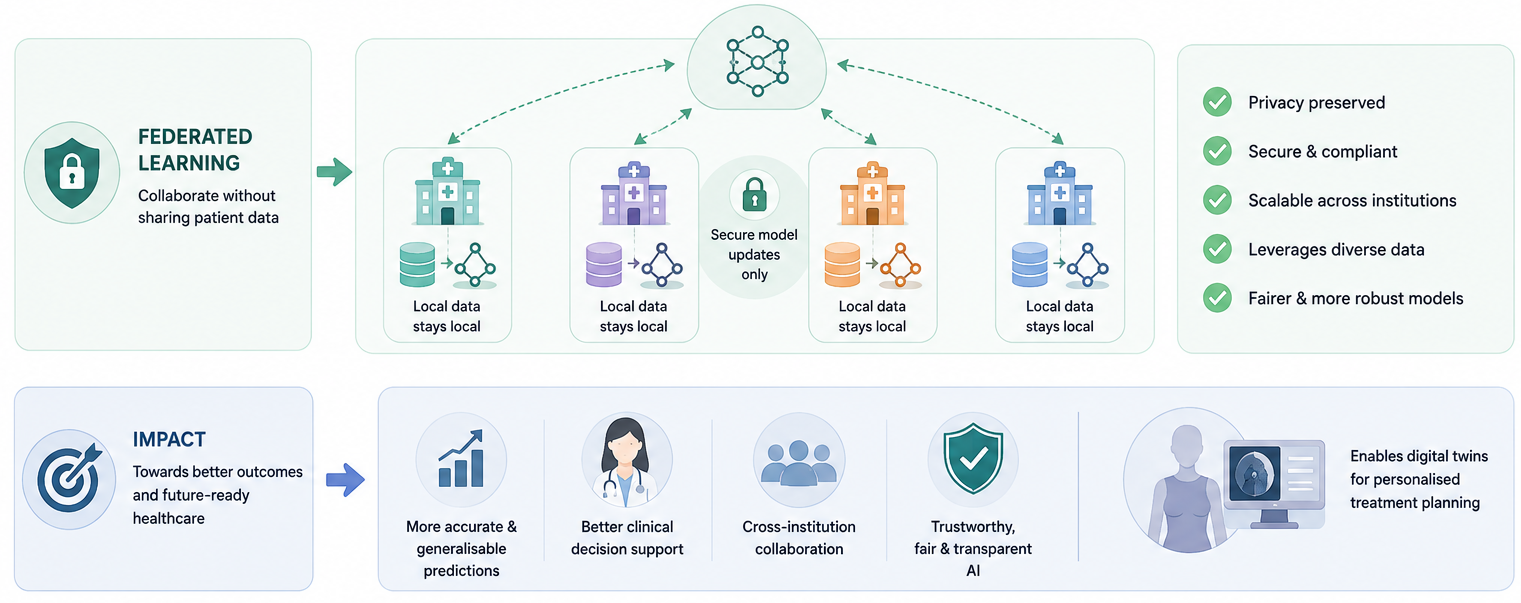}
\end{graphicalabstract}

\begin{highlights}
    \item Proposed a trustworthy multimodal federated learning framework for personalised breast cancer prediction.
    \item Integrated MRI, tumour segmentation masks and clinical data within a unified multimodal architecture.
    \item Demonstrated competitive performance against a centralised learning approach while preserving data privacy.
    \item Incorporated fairness-aware federated aggregation to improve equitable performance across patient groups.
    \item Explored the potential of federated learning to support future digital twin–enabled personalised cancer care.
\end{highlights}

%% Keywords

\end{frontmatter}




\section{Introduction}
Federated Learning (FL) was first introduced as a way for multiple institutions to train models together without sharing raw data \cite{mcmahan2017communication}. Since then, it has been widely explored in healthcare, particularly for medical imaging and electronic health records, where privacy regulations make data sharing difficult \cite{kaissis2020secure}. Although FL reduces the need to centralise patient data, research shows that it does not automatically solve issues related to security, fairness, transparency, or scalability. For example, model updates can still be vulnerable to attacks, and biased local datasets may lead to unfair global models. To address these concerns, researchers have proposed adding techniques such as differential privacy, homomorphic encryption, blockchain technology, and fairness-aware aggregation methods. However, most studies focus on improving only one aspect of FL rather than developing a complete system that could realistically be deployed in healthcare.

Personalised medicine requires large and diverse datasets, which are rarely available within a single hospital. While some research combines FL with personalised treatment modeling, there is still limited work examining how such systems can be made secure, scalable, and suitable for real-world use \cite{ciardiello2014delivering}.

A digital twin (DT) is a virtual model that mirrors a real-world entity. In a clinical context, this concept can be applied to individual patients, enabling the simulation of tumour progression and the modelling of treatment response over time. Current DT research often relies on mathematical modeling or imaging data, but few studies integrate DT development with federated AI systems across institutions. Many remain proof-of-concept and lack strong privacy-preserving infrastructures \cite{d2024recent}.

Few studies have examined how FL can be improved to address security, transparency, fairness, and scalability at the same time, while also enabling multimodal tumour prediction across institutions. In addition, limited research directly compares enhanced FL systems with centralised models to determine whether similar performance can be achieved \cite{mondal2024ai}. Addressing this gap is important for moving federated AI from theory into practical clinical use. Overall, research in this area will enhance patient outcomes, improve diagnostic accuracy, and deepen our understanding of tumour behaviour\cite{myrzashova2024safeguarding}.
\begin{figure}[!t]
    \centering
    \includegraphics[width=\linewidth]{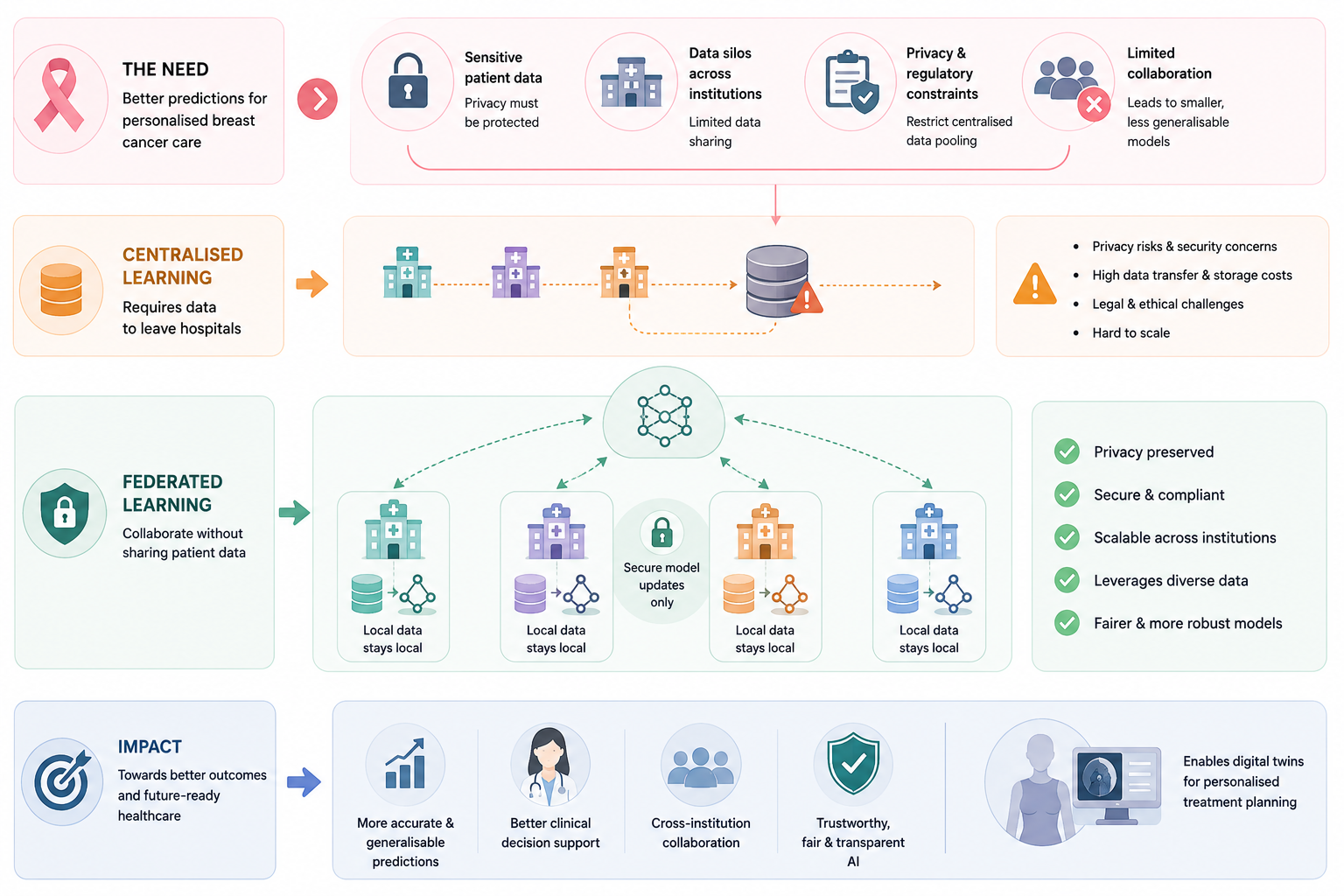}
    \caption{Motivation for federated learning in personalised breast cancer prediction, highlighting privacy-preserving collaboration across healthcare institutions to enable secure, scalable, and trustworthy AI-driven clinical decision support.}
    \label{fig:motivation}
\end{figure}

The Figure \ref{fig:motivation} illustrates the motivation for adopting federated learning in personalised breast cancer prediction. Traditional centralised approaches face privacy, regulatory, and data-sharing challenges, limiting collaboration across healthcare institutions. Federated learning enables secure collaborative model training while keeping patient data local, supporting privacy preservation, scalability, fairness, and trustworthy AI. Ultimately, it facilitates more accurate predictions, better clinical decision support, and future digital twin–enabled personalised treatment planning.

\section{Background}

A structured literature review was conducted across three primary themes: federated learning, personalised treatment, and digital twins. Relevant studies were identified through searches of PubMed, ScienceDirect, and IEEE Xplore.
 
 \subsection{Secure Federated Learning for Medical Data}

The challenges facing medical data management include the absence of standardised electronic health records (EHRs), the risk of patient data exposure, and the requirement to comply with regulatory frameworks. Although anonymisation and pseudonymisation were explored as privacy-preserving techniques, both have been deemed insufficient \cite{kaissis2020secure}. FL has emerged as a compelling alternative, as it enables model training without requiring raw data to leave the institution. However, FL alone does not guarantee security; model updates can be intercepted and partial data reconstruction from shared weights remains \cite{kaissis2020secure}.

To address this, several complementary security mechanisms are proposed. Differential privacy introduces statistical noise into model updates to prevent individual data points from being traced. One proof-of-concept study applied Gaussian noise to model weights \cite{jimenez2023memory}. Homomorphic encryption offers an alternative approach, allowing model updates to be \cite{myrzashova2024safeguarding}. Blockchain technology was also explored as a means of maintaining a transparent and tamper-proof record of model updates, with \cite{kuliha2024secure}. Two experimental studies demonstrated that combining blockchain technology with FL improved both overall performance and security, supporting the \cite{alzubi2022cloud, myrzashova2024safeguarding}. Notably, one study using Ethereum blockchain achieved \cite{myrzashova2024safeguarding}, while a separate study using consortium blockchain produced results closely comparable to a centralised \cite{9774951}. The key distinction between the two is that consortium blockchain operates on a permissioned basis, whereas Ethereum is publicly accessible.

A range of aggregation strategies were evaluated across the literature. FedYogi improved precision and recall by 9.91\% and 10.8\%, respectively, compared to alternative methods \cite{Yeom_Kim_Kim_Kim_Kim_2024}, while FedAvgM achieved a \cite{dang2022federated}. Architecturally, convolutional neural networks (CNNs) were the most commonly \cite{alzubi2022cloud, almufareh2023federated}. One study trained a Conv3DNet across four hospitals in China using the Flower framework and found that the FL model outperformed its centralised counterpart, achieving 75\% accuracy \cite{liu2024predicting}.

Fairness was also identified as an important consideration. A proof-of-concept study applied the Ditto framework to address group fairness by incorporating a generalisation term into local training, yielding fairer outcomes than standalone training \cite{wang2024analyzing}. A further study noted that local fairness does not necessarily translate to global fairness and proposed FedGFT, an aggregation method that uses local \cite{wang2023mitigating}. 

Regulatory compliance presents an additional challenge, particularly in cross-border settings where legislation such as GDPR restricts data sharing. Smart contracts have been proposed as a mechanism for encoding predefined data access and sharing rules within \cite{ali2023empowering}.

Collectively, the literature indicates that while FL represents a significant advantage over centralised approaches for sensitive medical data, it must be complemented by at least one additional security mechanism, whether differential privacy, homomorphic encryption, or blockchain technology, to be viable in real-world deployments \cite{kaissis2020secure}.

\subsection{Personalised Treatment Using Federated Learning}

Personalised treatment has the potential to substantially improve patient outcomes by tailoring clinical decisions to an individual's biological profile and circumstances, rather than a global base line. However, realising this potential requires large, well-annotated datasets that are regularly updated to account for tumour mutation and drug resistance \cite{ciardiello2014delivering}. One example of this approach in practice is Trastuzumab, a breast cancer therapy targeted specifically at patients who overexpress \cite{jackson2015personalised}.

Several studies have explored FL as a framework for delivering personalised treatment at scale. One experimental study combined FL, blockchain technology, and quantum computing to create a secure and scalable personalised medication system. Compared to a centralised AI baseline, this approach reduced latency by 350ms, increased security by 30\%, improved privacy preservation by 49\%, and reduced energy consumption by 200J \cite{mondal2024ai}. A prototype study using the Flower framework evaluated several model architectures, finding that BERT+LSTM achieved the strongest performance, reaching 98.75\% aggregated accuracy by the third federated round \cite{sehag2024federated}. A further study introduced MetaFed, a framework combining FL with cyclic knowledge distillation, in which clients share predictions rather than model weights. This approach consistently outperformed FedAvg, FedProx, and FedBN across multiple datasets, while also offering improved security and architectural flexibility \cite{chen2023metafed}.

The reviewed literature indicates that no single model architecture consistently delivers optimal results across datasets; model selection must therefore be guided by the characteristics of the data in question. A recurring concern is the computational demand associated with large personalised datasets, which risks introducing demographic bias where institutions with limited resources are underrepresented in the global model.
% \begin{figure}[!b]
%     \centering
%     \includegraphics[width=0.7\linewidth]{Images/wordcloud_BC.png}
%     \caption{Enter Caption}
%     \label{fig:placeholder}
% \end{figure}

\subsection{Personalised Treatment Through Digital Twins}

Digital twins are virtual replicas of real world entities that use real time data to simulate and predict the behaviour of their physical counterpart \cite{mollica2024digital}. In the clinical domain, DTs offer the potential not only to personalise treatment but also to support virtual drug trials \cite{checcucci2024digital}. Key barriers to their adoption include biological complexity, clinical usability, regulatory compliance, and the scarcity of clinical data \cite{li2025digital}. Synthetic data generation has been proposed as one approach to partially address the data availability problem \cite{padoan2024dynamic}.

Mathematical modelling forms the foundation of most tumour-based DT systems. Capturing tumour behaviour requires models that can simulate cell proliferation \cite{d2024recent}. One study applied a reaction-diffusion partial differential equation to model tumour evolution in response to neoadjuvant chemotherapy, achieving an AUC of 89\% for both total tumour cellularity and total tumour volume across fifty patients \cite{wu2022mri}. A separate study modelled prostate tumour growth using thermoelastic expansion equations, reporting a relative growth error of between 2.56\% \cite{perez2024patient}. 

Despite the breadth of research into mathematical tumour modelling, comparatively little work has integrated AI to personalise these models at the individual patient level. Incorporating patient-specific factors such as genetics and lifestyle into DT frameworks could improve treatment selection and help avoid ineffective therapies, with downstream benefits for both patient outcomes and healthcare costs.

Across all three themes, a consistent gap emerges in the literature: security, scalability, fairness, and transparency are typically studied in isolation rather than as components of a unified system. This study seeks to address that gap by examining how these elements can be integrated within a single FL framework, and by exploring the feasibility of using multimodal data to move toward personalised, DT-informed cancer treatment planning.

%%%%%%%%%%%%%%%%%%%%%%%%%%%%%%%%%%%%%%%%%%
\begin{figure}[!t] 
\centering    
\includegraphics[width=1.0\textwidth]{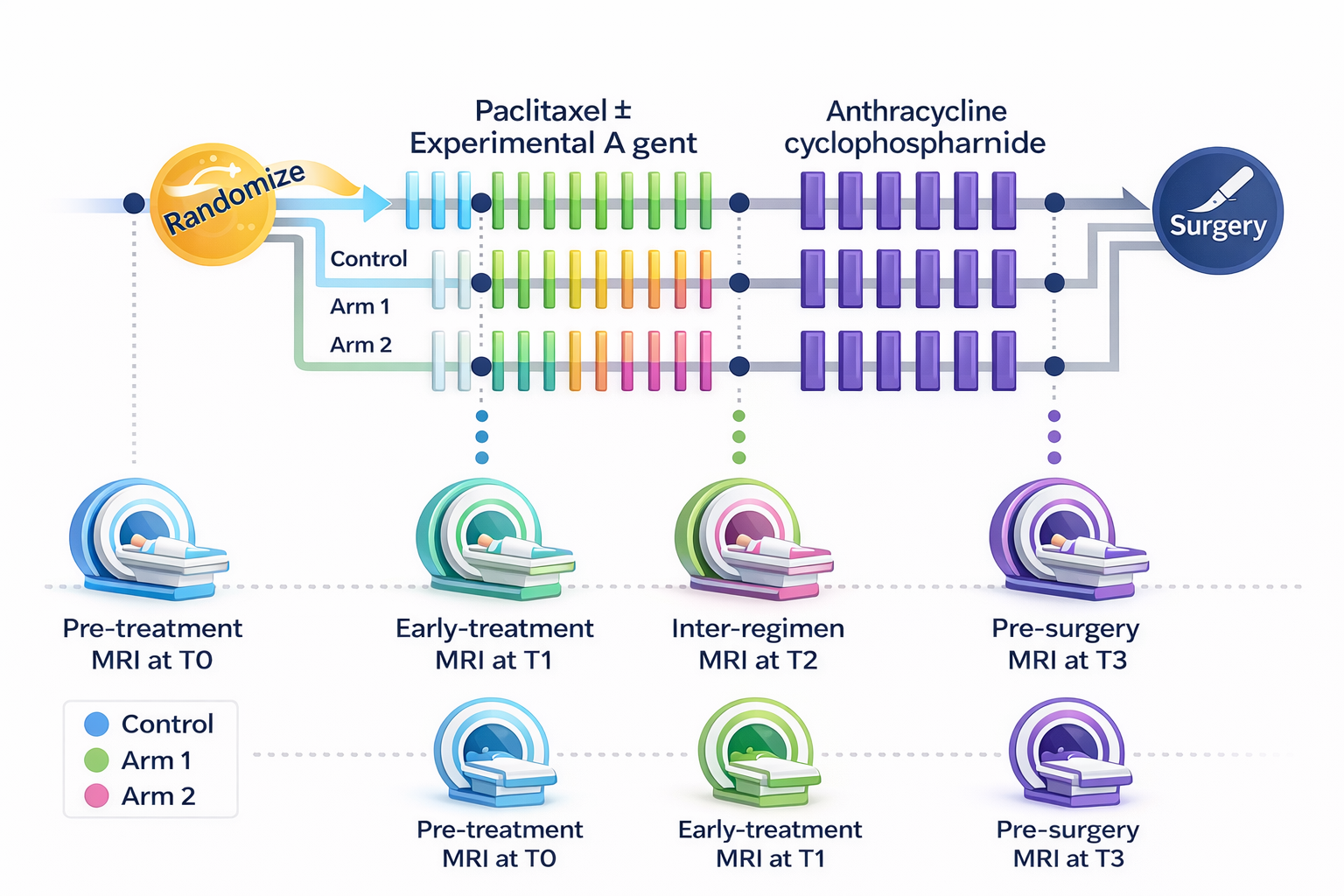}
\caption{The image illustrates the data collection timeline. T0 represents the baseline before any treatment, T1 involves the administration of various experimental treatments, T2 marks the point at which all patients receive Anthracycline Cyclophosphamide, and the final time point occurs just before surgery, following treatment \cite{Li2022ISPY2}.}
\label{fig:data}
\end{figure}

\section{Materials and Methods}
% \subsection{Study Design}
This research investigates whether an FL framework can be used to predict tumour progression in breast cancer patients using multimodal data while addressing four key deployment pillars: scalability, transparency, security, and fairness. The performance of the federated model was compared directly with a centralised model trained on pooled data to evaluate whether similar predictive accuracy can be achieved without the risk of sharing raw patient data.
Only components that directly influenced measurable outcomes, such as predictive performance and fairness, were implemented experimentally. Security mechanisms including blockchain integration and differential privacy are described conceptually but were not experimentally evaluated due to infrastructure constraints.

\subsection{Dataset}

The dataset used in this study was the I-SPY2 breast cancer dataset obtained from The Cancer Imaging Archive (TCIA) \cite{Li2022ISPY2}. The dataset is publicly available for research purposes and contains de-identified patient data. As no new patient data was collected and all data was anonymised prior to release, additional ethical approval was not required for this study.

The dataset includes multimodal information consisting of tabular clinical and demographic variables, MRI scans, and corresponding region-of-interest (ROI) masks. Data was collected across four time points (T0–T3). In this work, data from T0 to T2 were used as inputs to predict tumour characteristics at T3, as shown in Figure~\ref{fig:data}.

\subsection{Data Preprocessing}

\subsubsection{Tabular Data Processing}
\begin{figure}[!b] 
\centering    
\includegraphics[width=1.0\textwidth]{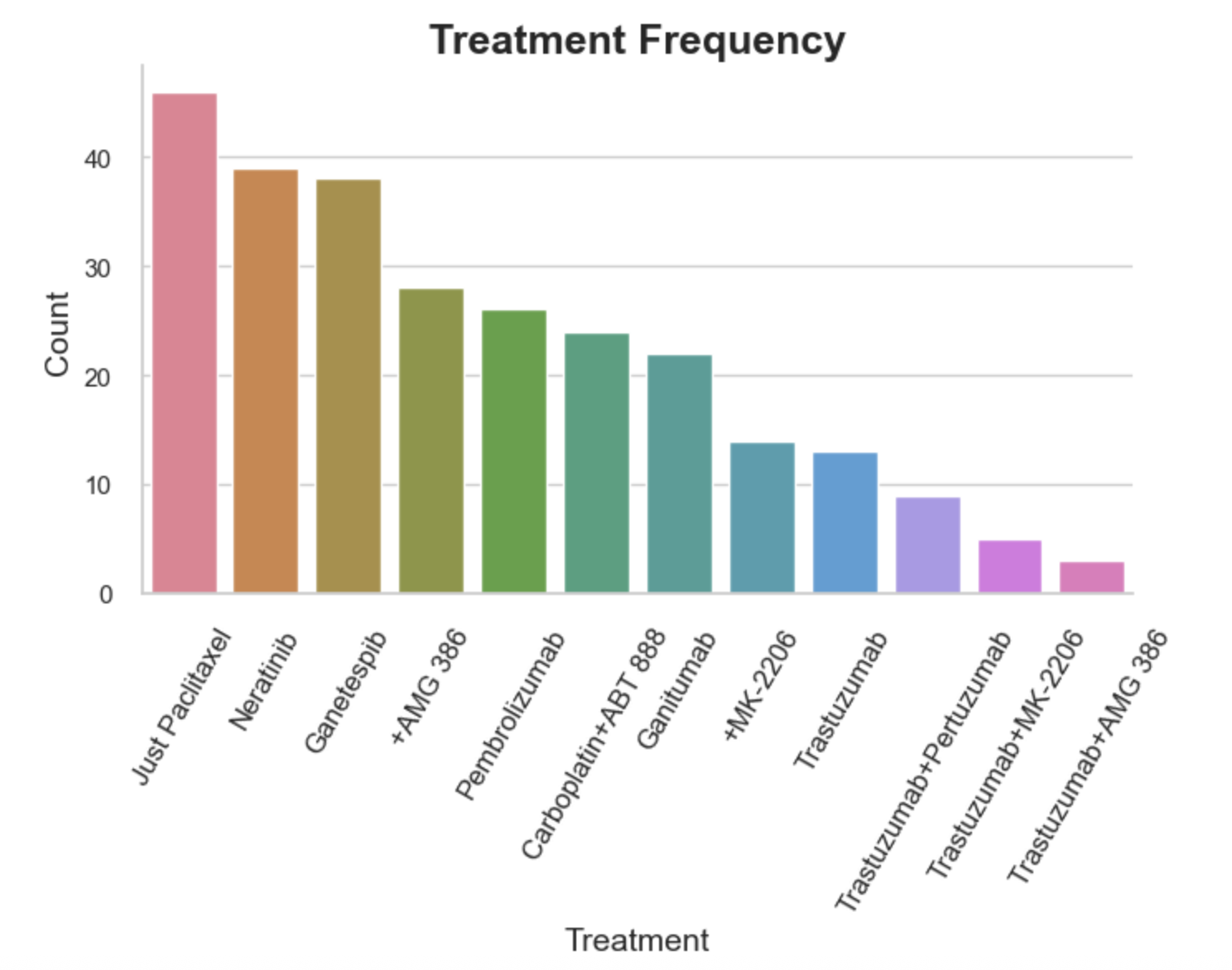}
\caption{The bar chart illustrates the distribution of combined treatments received by patients in addition to Paclitaxel.}
\label{fig:freq}
\end{figure}

The tabular dataset was restructured so that each patient had four rows corresponding to the four time points. Static features, such as menopausal status, were duplicated across time points to preserve complete patient information. Dates were converted to the number of days to retain temporal structure while removing irrelevant calendar information. Input features included tumour volume, sphericity, longest diameter, hormone receptor status, HER2 status, MP, pCR, age at screening, length of treatment, race, menopausal status, and treatment type. Target variables at T3 included tumour volume, sphericity, longest diameter, MRI volume, and ROI mask. Categorical variables were one-hot encoded, and numerical features were scaled using Min–Max normalisation. The treatment “Anthracycline cyclophosphamide” was removed because it was administered to all patients and therefore provided no value. The patients treatment combinations were consolidated into a single categorical variable as shown in Figure~\ref{fig:freq}.

\subsubsection{Imaging Data Processing}
\begin{figure}[!t] 
\centering    
\includegraphics[width=1.0\textwidth]{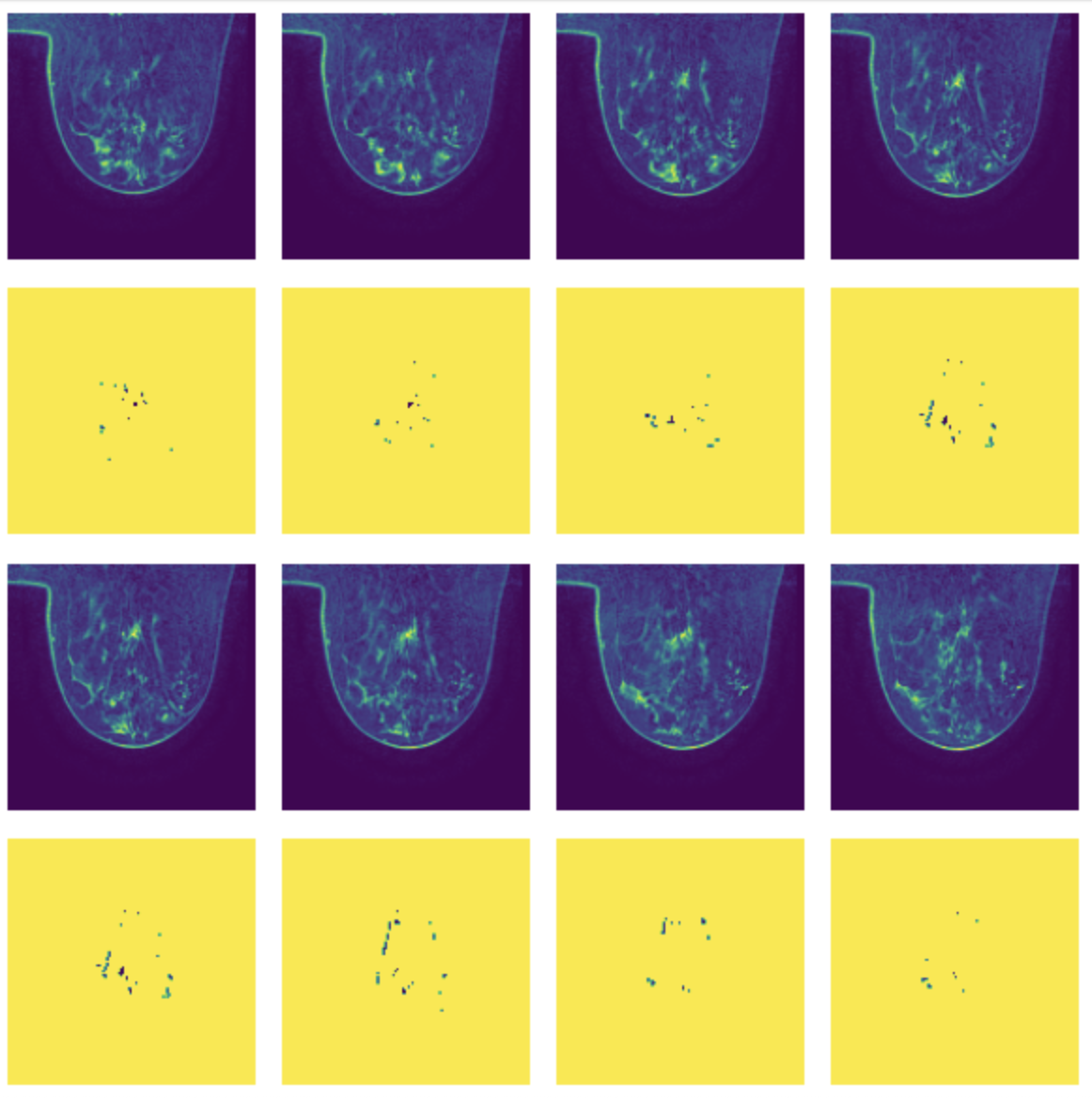}
\caption{The figure presents the 8 slices used for one patient where it can be seen that there is little movement between one slice to the next.}
\label{fig:data_images}
\end{figure}

MRI and ROI images were resized to 128 × 128 pixels to reduce computational load. Eight slices per time point were selected due to memory limitations. Three time points (T0–T2) were used as input. MRI pixel values were normalised by dividing by 255 to ensure consistency across modalities. MRI and ROI volumes were concatenated along a new channel dimension, producing a tensor of shape (3, 8, 128, 128, 2), where three represents time points, eight represents slices per time point, 128 × 128 represents spatial resolution, and two represents MRI and ROI modalities.
Although some patients had up to 800 ROI slices, full volume training was not feasible within the available hardware constraints. This reduction enabled parallel federated training across clients while maintaining temporal and spatial tumour information. Figure \ref{fig:data_images} shows the eight slices selected for one patient's appointment, where we can see that the ROI varies slightly across slices. Each MRI pixel was then normalised by dividing its value by 255 to ensure consistency in scale with the ROI and associated tabular data.

\subsection{Federated Learning Framework}
Federated learning was implemented using the Flower framework following the decentralised training approach introduced by \cite{mcmahan2017communication}. Each client trained the model locally on its own partitioned dataset and transmitted model parameters to a central server for aggregation.
\begin{figure}[!b] 
\centering    
\includegraphics[width=1.0\textwidth]{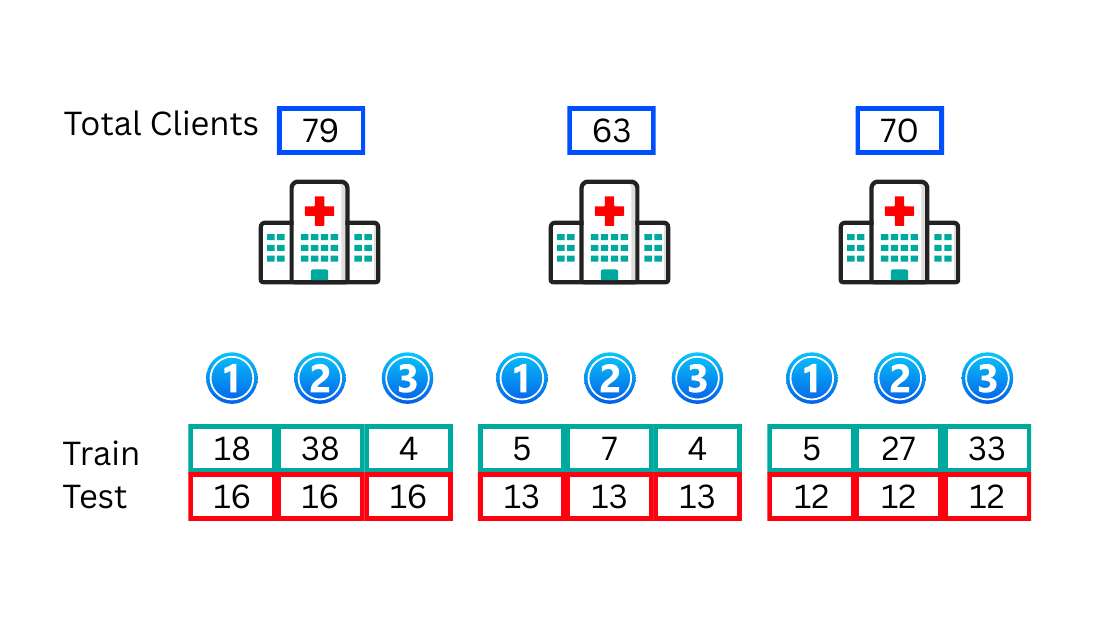}
\caption{The figure shows how the original data was uniformly partitioned between clients and then how the data was direchlet partitioned between rounds for each client. }
\label{fig:partition}
\end{figure}

\subsection{Experimental Setup}
The system was deployed on a single laptop environment consisting of one global server and a virtual machine hosting three clients. Docker was used to containerise each component to ensure reproducibility and isolate dependencies. Each client operated within its own container and accessed a distinct dataset partition, simulating multi-institutional training while maintaining computational efficiency. Although a hierarchical federated structure incorporating supernodes and blockchain integration was conceptually designed, the experimental evaluation was conducted using a simplified architecture due to TLS certificate limitations within the Flower framework and hardware constraints. Both components would meaningfully strengthen a real-world deployment. Supernodes improve scalability by introducing intermediate aggregation stages, distributing the computational load that would otherwise bottleneck a single global server. Blockchain integration would enhance security and provide a transparent audit trail of all transactions across participating institutions.

\begin{figure}[!b]   
\includegraphics[width=1.0\textwidth]{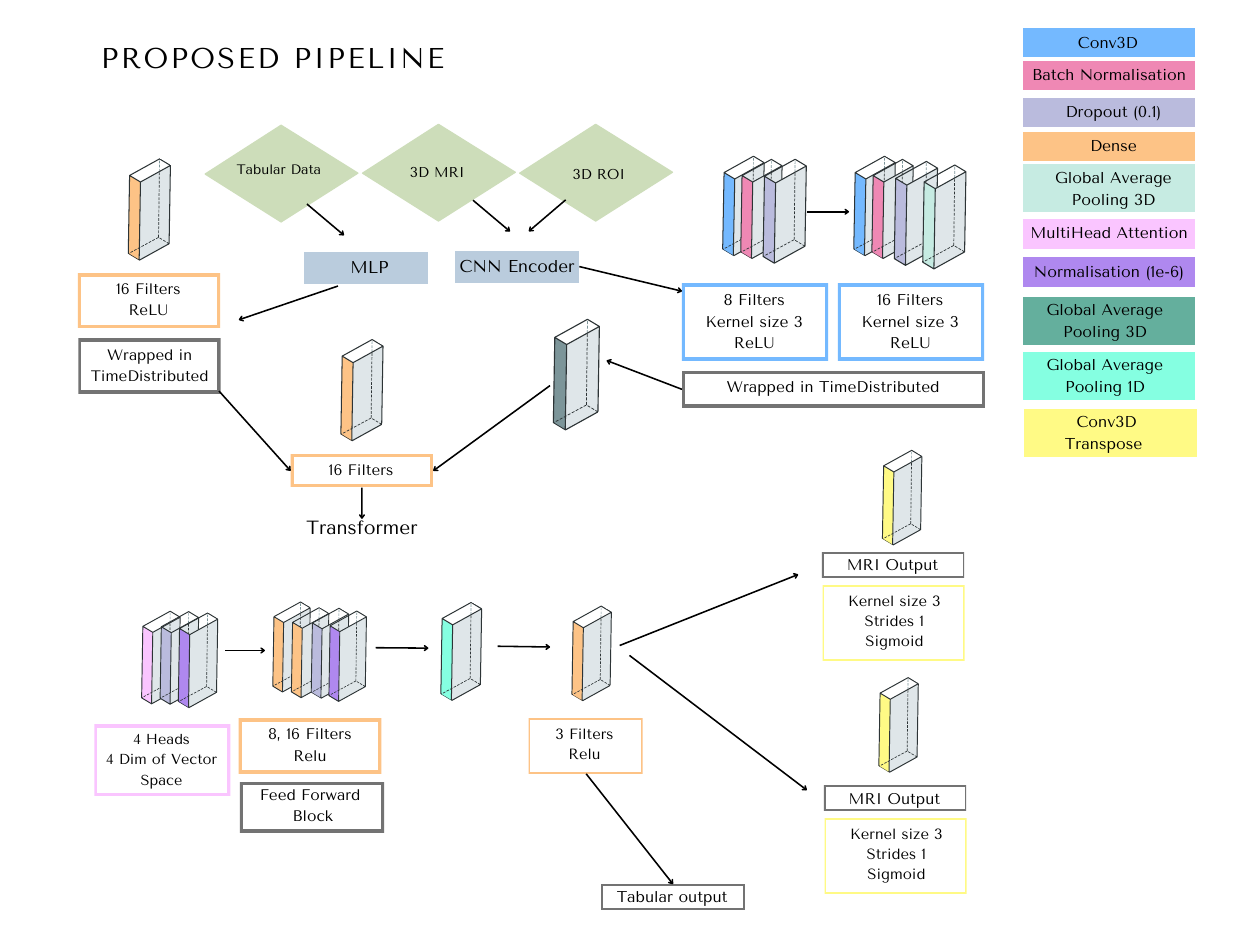}
\caption{The pipeline illustrates how the tabular and image data were processed separately, each undergoing transformations to create inputs which then can be concatenated together, so a single input can be used for the model, which then outputs all three modes individually.}
\label{fig:model_pipeline}
\end{figure}

\subsection{Data Partitioning}
Twenty-five patients were reserved as a centralised global test set. The remaining patients were distributed across three clients. An initial uniform split was performed to split between the clients, followed by Dirichlet partitioning during training rounds to simulate non-identical data distributions across institutions. Test sets remained fixed across training rounds to ensure consistent evaluation of model performance over time. (Figure~\ref{fig:partition}).

\subsection{Model Architecture}
The multimodal model consisted of an image encoder, a tabular encoder, and a transformer-based fusion module which can be viewed in Figure~\ref{fig:model_pipeline}. The image encoder was a 3D convolutional neural network containing two convolutional layers with 8 and 16 filters respectively, followed by 3D global average pooling to reduce dimensionality.
The tabular encoder consisted of a single dense layer with 16 units.

The resulting feature vectors were concatenated and passed through a dense layer of size 16 before entering a transformer block inspired by \cite{Vaswani_Shazeer_Parmar_Uszkoreit_Jones_Gomez_Kaiser_Polosukhin_2023}. The transformer used four attention heads, a dropout rate of 0.1, and layer normalisation with epsilon set to 1e-6. The feedforward network consisted of dense layers of sizes 8 and 16. A one-dimensional average pooling layer was applied before final output processing. Three outputs were generated. Tabular predictions for T3 tumour measurements were produced through a dense layer with three units. MRI and ROI outputs were reconstructed using Conv3DTranspose layers with kernel size 3 and stride 1, and a sigmoid activation function to constrain output values between 0 and 1. Mean Absolute Error was used as the loss function for tabular and MRI regression tasks. Binary cross-entropy was used for ROI mask prediction. Accuracy was used as the evaluation metric for segmentation performance.

\emph{Federated Aggregation:}
Federated Averaging (FedAvg) was used as the baseline aggregation method:
\begin{equation}
     w=\sum{^K_{k=1}\frac{n_k}{n}w_k}
\end{equation}
where $K$ is the number of clients, $n_k$ is the amount of data for client $k$, $n$ is the amount of data across all clients and $w$ is the weights.

\emph{Fairness Evaluation:} To evaluate demographic fairness, race was selected due to its uneven distribution within the dataset. Federated Group Fairness Training (FedGFT) was implemented to prevent unfair results \cite{wang2023mitigating}. Each client transmitted the number of samples per racial group and the corresponding mean absolute error for each group. The server computed a fairness penalty defined as the difference between the maximum and minimum MAE across groups. This penalty was incorporated into optimisation using the formulation:
\begin{equation}
    \frac{1}{1+penalty}
\end{equation}

\emph{Scalability and Reproducibility:}
Docker was used to define a static and reproducible training environment. Each training run generated three files: $results.json$, $run-config.json$, and $final-model.weights.h5.$ These were stored in time-stamped directories to ensure traceability and allow full replication of experimental configurations. A centralised baseline model was trained using pooled data and identical architecture to provide a direct comparison with federated training.

\emph{Security Considerations:} Blockchain integration, differential privacy, and TLS certificate encryption were examined conceptually but were not experimentally implemented. Integration of TensorFlow Privacy proved incompatible with the Docker environment at the time of experimentation, preventing its evaluation. Additionally, as all experiments were conducted on a single machine, it was not possible to meaningfully simulate or assess real-world security threats. These mechanisms are therefore discussed as proposed enhancements for future deployment rather than experimentally validated components.

\emph{Availability of Code and Materials:} All preprocessing scripts, federated training code, and model architecture definitions will be made publicly available in a repository upon publication. The dataset is publicly accessible through The Cancer Imaging Archive\cite{Li2022ISPY2}.

\emph{Generative AI Disclosure:} Generative artificial intelligence tools were used to assist with language refinement and structural editing of the manuscript. AI tools were not used for data generation, model training, experimental design, or statistical analysis.

%%%%%%%%%%%%%%%%%%%%%%%%%%%%%%%%%%%%%%%%%%
\section{Results}
To establish a performance baseline and validate the FL framework, a centralised model was first trained using the complete global dataset. All subsequent FL models were evaluated against this baseline using the same global test set to ensure a consistent and fair comparison. Given the experimental constraints specifically, the use of only eight low-resolution MRI slices and all clients operating on a single machine with limited computational resources, absolute performance values should be interpreted accordingly. The primary objective is to assess whether FL can serve as a viable alternative to centralised training, rather than to achieve state of the art performance.

%-----------------------------------------------------------
\subsection{Centralised vs.\ Federated Learning Performance}
%-----------------------------------------------------------

Table~\ref{tab:comp_table} presents loss and task specific performance metrics across the centralised model and four rounds of FL training. FedGFT was disabled for this comparison to isolate the effect of federated aggregation alone.

% \begin{table}[h]
% \caption{Caption text}\label{tab1}%
% \begin{tabular}{@{}llll@{}}
% \toprule
% Column 1 & Column 2  & Column 3 & Column 4\\
% \midrule
% row 1    & data 1   & data 2  & data 3  \\
% row 2    & data 4   & data 5\footnotemark[1]  & data 6  \\
% row 3    & data 7   & data 8  & data 9\footnotemark[2]  \\
% \botrule
% \end{tabular}
% \footnotetext{Source: This is an example of table footnote. This is an example of table footnote.}
% \footnotetext[1]{Example for a first table footnote. This is an example of table footnote.}
% \footnotetext[2]{Example for a second table footnote. This is an example of table footnote.}
% \end{table}

% \noindent

\begin{table}[!t]
\centering
\label{tab:comp_table}

\begin{tabular}{@{}lccccc@{}}
\toprule
\textbf{Metric} & \textbf{Centralised} & \textbf{FL R0} & \textbf{FL R1} & \textbf{FL R2} & \textbf{FL R3} \\
\midrule
Loss         & 3.7253 & 1.8214 & 1.2707 & 0.8923 & 0.7477 \\
Tabular MAE  & 0.2029 & 0.6128 & 0.2922 & 0.1754 & 0.2072 \\
MRI MAE      & 0.4186 & 0.5612 & 0.4704 & 0.3731 & 0.3420 \\
ROI Accuracy & 0.9137 & 0.7651 & 0.9847 & 0.9926 & 0.9995 \\
\midrule
% \botrule
\end{tabular}
\caption{Comparison of loss and performance metrics between the centralised model and the FL model across three training rounds.}
\end{table}

\begin{figure}[!t] 
\centering    
\includegraphics[width=1.0\textwidth]{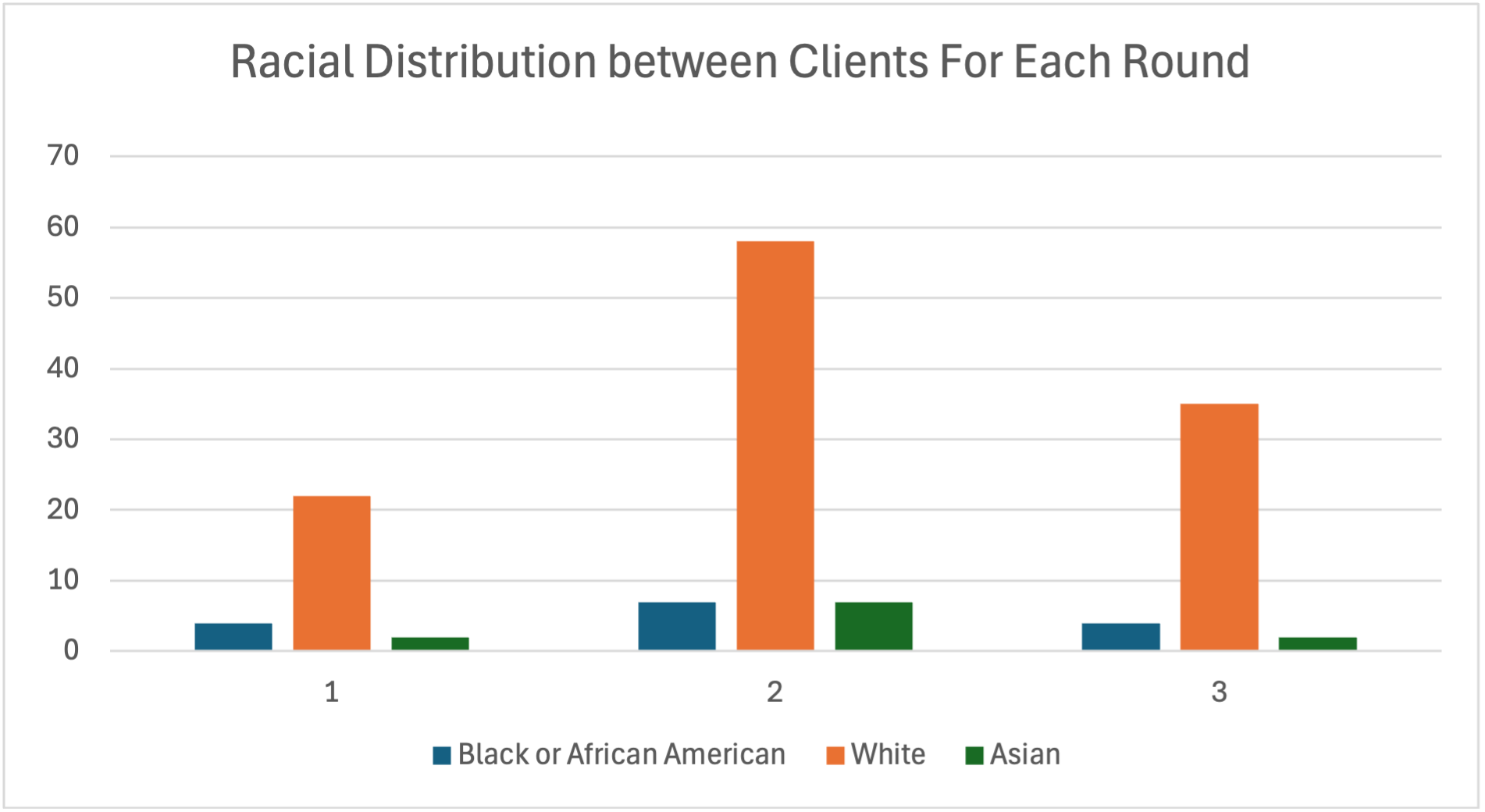}
\caption{The bar chart presents the distribution of different races used within each partition for each round when using FedGFT.}
\label{fig:fairnesskew}
\end{figure}

At initialisation (R0), the FL model underperforms the centralised baseline across all metrics, which is expected prior to any local training. By Round~1, the FL model achieves comparable tabular MAE (0.2922 vs.\ 0.2029) and MRI MAE (0.4704 vs.\ 0.4186), while substantially exceeding the centralised model on ROI accuracy (0.9847 vs.\ 0.9137). By Round~3, the FL model surpasses the centralised baseline across every reported metric, achieving a loss of 0.7477, tabular MAE of 0.2072, MRI MAE of 0.3420, and ROI accuracy of 0.9995. These results demonstrate that, even within a constrained experimental setup, federated learning represents a viable and potentially superior alternative to centralised training.

%-----------------------------------------------------------
\subsection{FedAVG vs.\ FedGFT Aggregation}
%-----------------------------------------------------------

A second experiment compared the standard FedAvg aggregation strategy against FedGFT, a fairness-aware variant that penalises models exhibiting biased performance across demographic subgroups. Results at Round~3 are reported in Table~\ref{tab:fairness_table}.

\begin{table}[h]
\centering

\label{tab:fairness_table}

\begin{tabular}{@{}lcc@{}}
\toprule
\textbf{Metric} & \textbf{FedAVG} & \textbf{FedGFT} \\
\midrule
Loss         & 0.7477 & 1.0040 \\
Tabular MAE  & 0.2072 & 0.1534 \\
MRI MAE      & 0.3421 & 0.2693 \\
ROI Accuracy & 0.9995 & 0.9972 \\
\midrule
\end{tabular}
\caption{Comparison of FedAVG and FedGFT aggregation strategies at Round~3.}
\end{table}

FedGFT underperforms in overall loss (1.004 vs.\ 0.7477) relative to FedAvg, but improves both MAE metrics and maintains competitive ROI accuracy (0.9972 vs.\ 0.9995). The dataset exhibits a demographic skew, with white patients being over represented relative to Black or African American and Asian patients across all client partitions (Figure~\ref{fig:fairnesskew}). FedGFT applied fairness penalties of 0.008619, 0.035103, and 0.068125 across Rounds~1, 2, and~3 respectively. The increasing penalty magnitude reflects growing sensitivity to inter-group performance disparities as training progresses, yet the penalties remain small in absolute terms, indicating that bias correction is achieved without materially degrading global model accuracy. Based on these results, FedGFT was adopted for all subsequent experiments.

%-----------------------------------------------------------
\subsection{Qualitative Evaluation of Image Segmentation}
%-----------------------------------------------------------

Despite the high ROI accuracy values reported in Table~\ref{tab:comp_table}, visual inspection of segmentation outputs reveals important limitations not captured by quantitative metrics alone. Where the vast majority of pixels correspond to background tissue, a model can attain near perfect pixel accuracy by predominantly predicting the non-tumour class. This is evident in the present results: 99\% ROI accuracy at Round~3 does not correspond to accurate tumour localisation.

\begin{figure}[!t] 
\centering    
\includegraphics[width=0.9\textwidth]{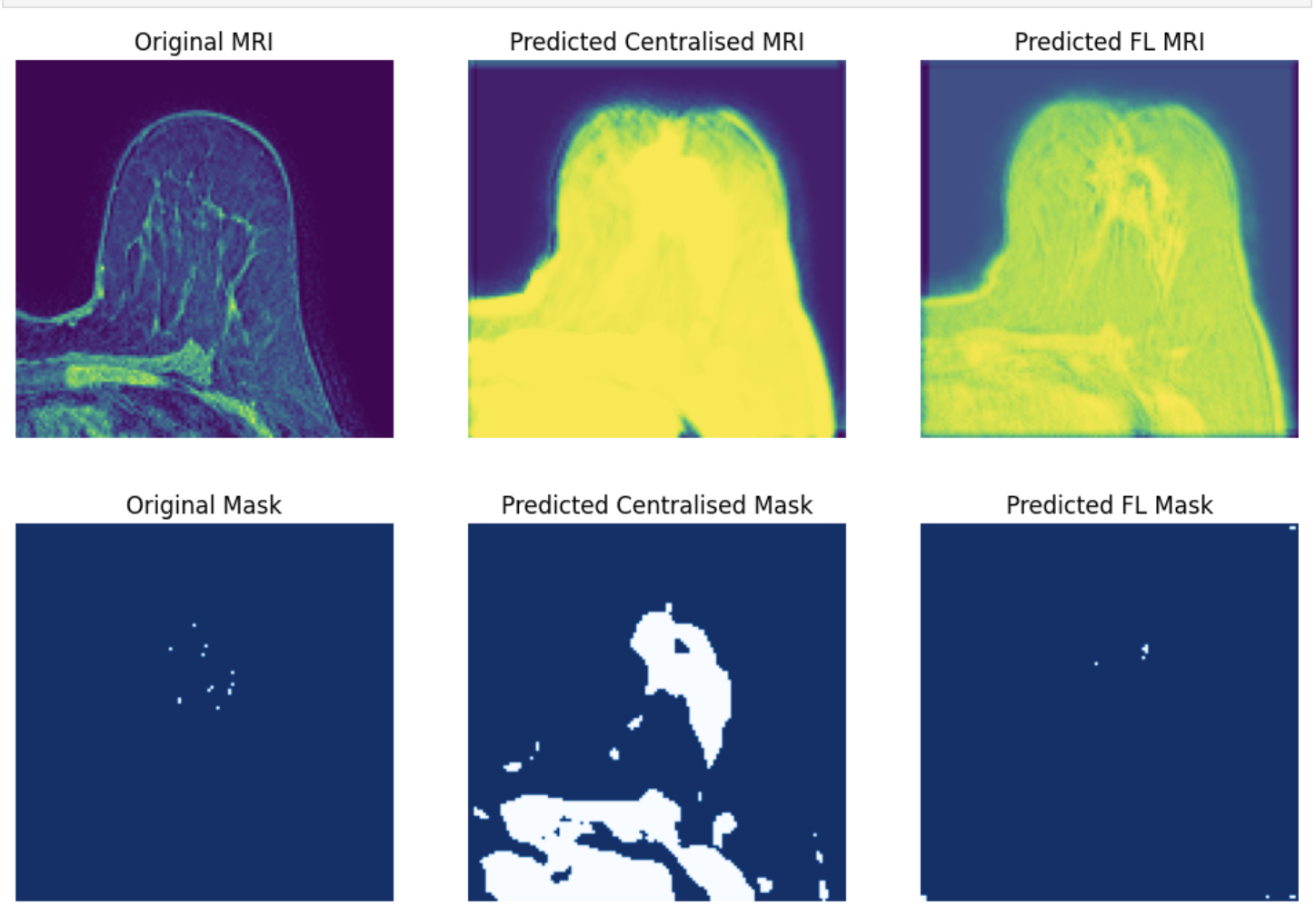}
\caption{This comparison of original and predicted MRI and mask images demonstrates that the centralised model significantly overestimates the mask region, whereas the FL model exhibits a tendency to underestimate the ROI.}
\label{fig:mrivsmask}
\end{figure}

\begin{figure}[!t] 
\centering    
\includegraphics[width=0.9\textwidth]{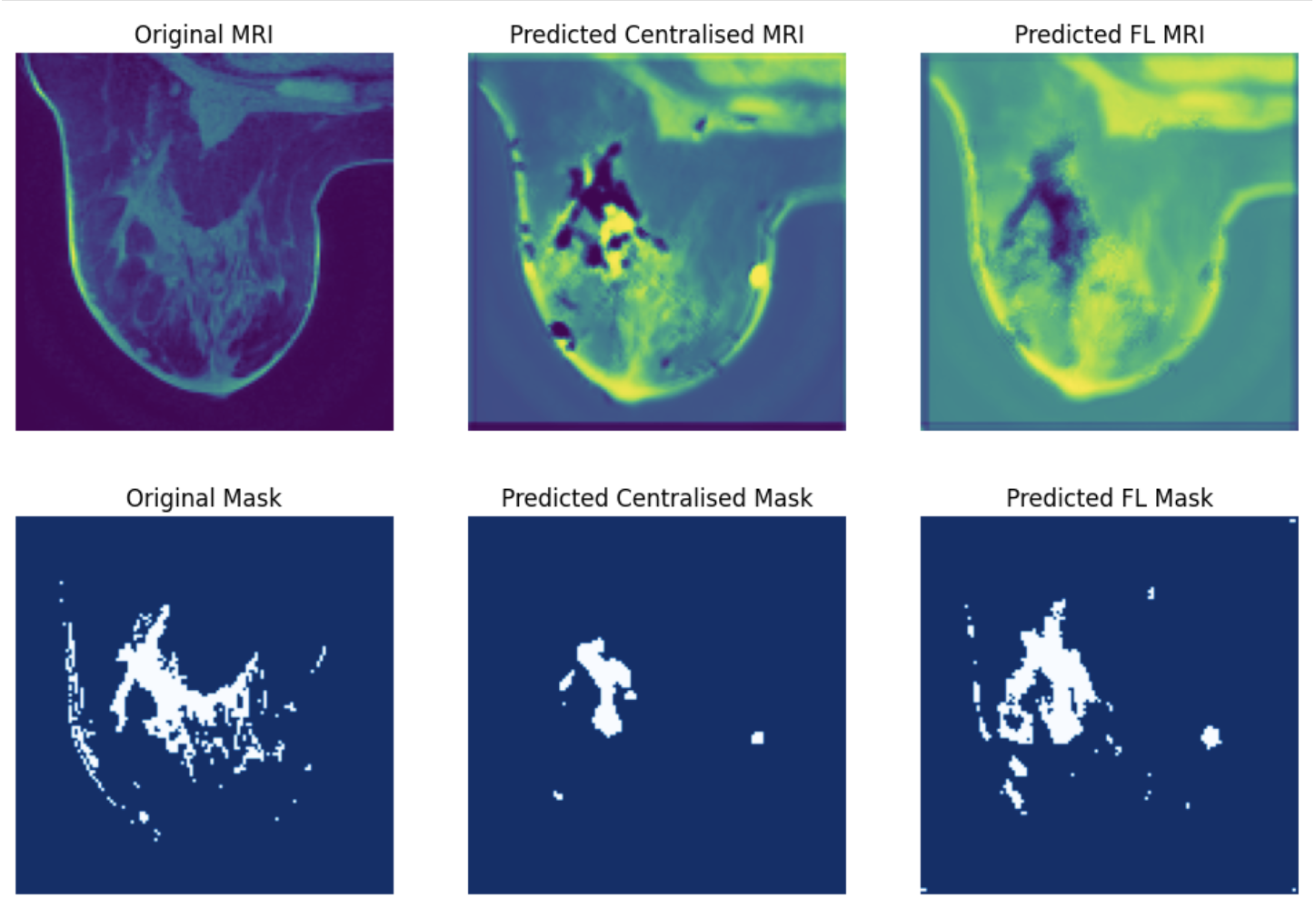}
\caption{A comparison of original and predicted MRI and mask images indicating that the FL model generates a relatively accurate mask, whereas the centralised model underestimates the complete mask region in this instance.}
\label{fig:mrivsmask2}
\end{figure}

Figures~\ref{fig:mrivsmask} and~\ref{fig:mrivsmask2} reveal the inaccuracies in the predicted mask. The centralised model overestimates the tumour, producing masks that are larger than the ground truth and spatially displaced from the true tumour location. The FL model, by contrast, tends to underestimate the ROI, identifying the approximate tumour region but capturing only a portion of its total area. In a second example, the FL model produces a more spatially accurate mask while the centralised model fails to predict the full ROI. Both models reconstruct MRI images lacking fine structural detail, with a tendency to enlarge the breast region. This is likely due to the the low input resolution and limited training set size. 
\begin{figure}[!t] 
\centering    
\includegraphics[width=0.9\textwidth]{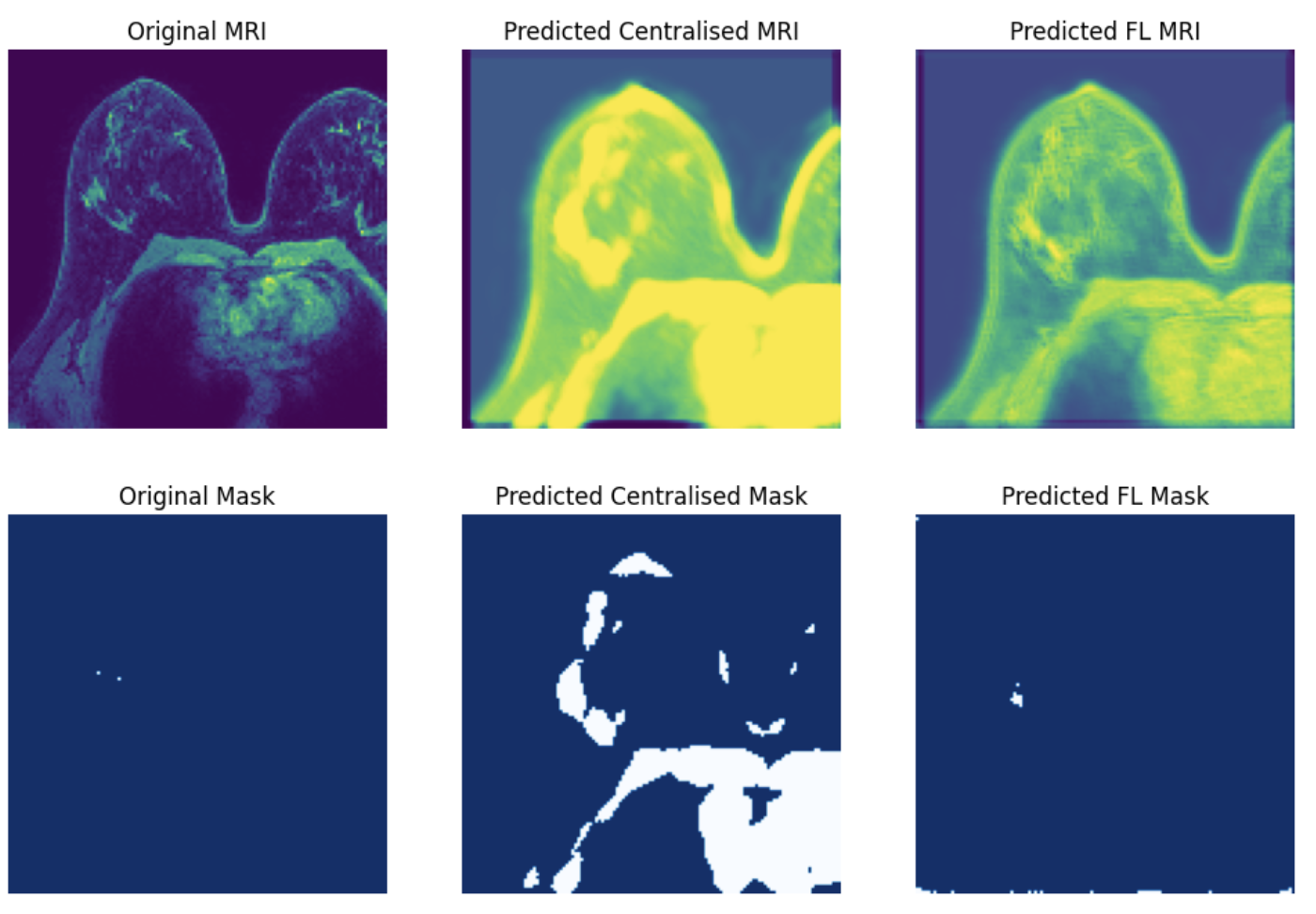}
\caption{This comparison compares the difference between a patient from one of the minority groups specifically Black or African American.}
\label{fig:minority}
\end{figure}

%-----------------------------------------------------------
\subsection{Minority Group Performance and Fairness}
%-----------------------------------------------------------
To assess the fairness of the FL model, predictions were examined for patients classified as Black or African American, a group substantially underrepresented in the training partitions. As shown in Figure~\ref{fig:minority}, the FL model with FedGFT produces outputs for minority patients that are qualitatively comparable to those obtained for the majority group. The centralised model again produces substantially over predicted segmentation masks for these patients, whereas the FL model more accurately approximates the ground truth tumour region. These findings suggest that the fairness weighted aggregation applied by FedGFT effectively mitigates the performance disparity that would otherwise arise from demographic imbalance in the training data, supporting its suitability for clinical applications across patient subgroups is a key requirement in order to obtain fair results.

\begin{figure}[!t] 
\centering    
\includegraphics[width=1.0\textwidth]{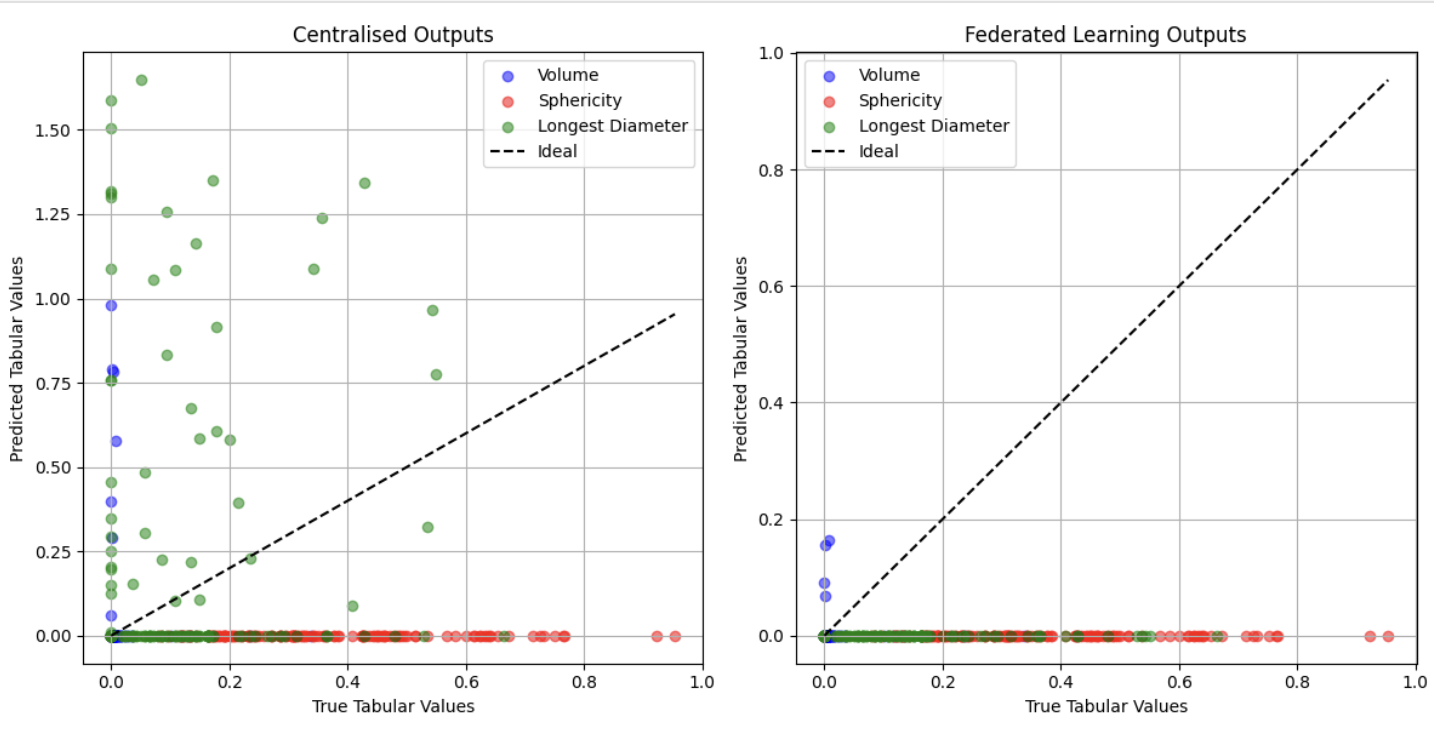}
\caption{An inspection of volume, sphericity, and longest diameter from the tabular outputs for both centralised and FL models reveals underfitting, with both models frequently predicting a value of zero for each feature.}
\label{fig:tab_comp}
\end{figure}

\begin{figure}[!t] 
\centering    
\includegraphics[width=1.0\textwidth]{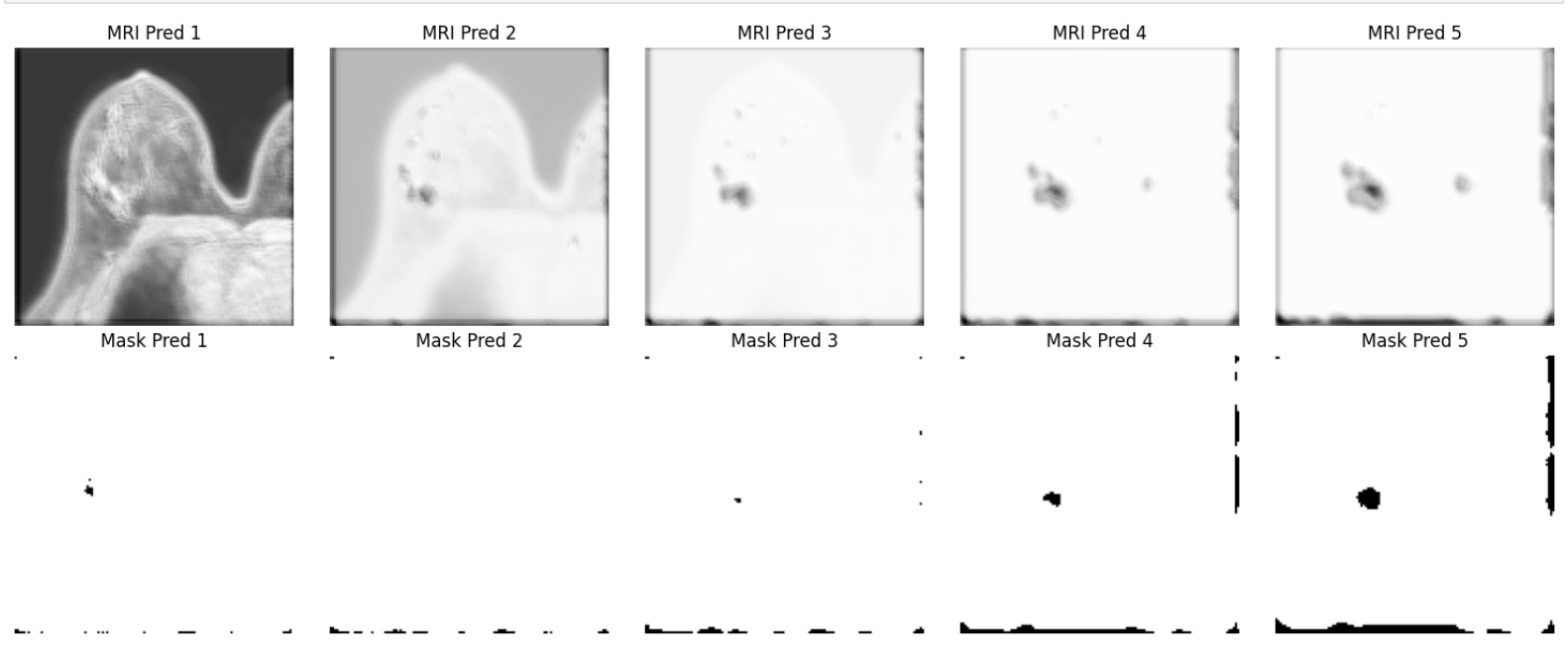}
\caption{The images present future predictions using the updated prediction as the next time step.}
\label{fig:future}
\end{figure}

%-----------------------------------------------------------
\subsection{Tabular Outcome Prediction}
%-----------------------------------------------------------

Both models exhibit substantial underfitting on the tabular prediction task, as illustrated by scatter plots of predicted versus true values for tumour volume, sphericity, and longest diameter (Figure~\ref{fig:tab_comp}). The diagonal $y = x$ line represents ideal prediction; neither model achieves alignment with this target. A marked tendency toward predicting zero is observed in both models,  particularly  in the FL model, where nearly all tabular outputs collapse to zero. 

The centralised model performs marginally better in tabular prediction, with a small number of outputs aligning with true values for the longest diameter; however, neither model produces meaningful predictions for sphericity or volume. This is a clinically significant limitation, as inaccurate tabular outputs could cause the model to underestimate the severity of a patient's condition, potentially leading to delayed or inappropriate treatment decisions. Developing richer feature representations and training on substantially larger datasets are therefore important priorities for future work.

%-----------------------------------------------------------
\subsection{Digital Twin Exploration}
%-----------------------------------------------------------

A preliminary exploration of the model's capacity for forecasting tumour devlopment was conducted to assess its feasibility as a DT component. The dataset provides four temporal imaging time points, with three used as input to predict the fourth. By treating the model's prediction at time step four as a new input, appending it to the temporal sequence while discarding the earliest observation, the system can generate a forecast for time step five, and iteratively beyond (Figure~\ref{fig:future}).

Applying this approach, the model predicts that by time step five there will be partial recurrence of the tumour. While the accuracy of these projections cannot be validated with the available data, the result demonstrates the conceptual viability of an iterative DT architecture. In a clinical deployment, regular imaging, for example weekly or monthly scans, could serve as a continuous input stream, enabling the DT to be a viable future which has great impact. However, while the potential clinical benefits of such a system are considerable, achieving results of a standard suitable for real-world deployment would require substantial computational resources and high quality, large scale datasets that may not be readily available across all participating institutions.

\begin{table}[!t]
\centering

\label{tab:comparison}
\renewcommand{\arraystretch}{1.3}
\footnotesize
\begin{tabular}{p{6.8cm}cccccc}
\toprule
\textbf{Capability} &
\textbf{Typical FL Studies} &
\textbf{This Work} \\
\midrule

Privacy-preserving collaborative learning &
\checkmark &
\checkmark \\

Multimodal data integration (clinical + MRI + ROI) &
Partial &
\checkmark \\

Comparison with centralised learning &
Limited &
\checkmark \\

Fairness-aware federated aggregation &
Limited &
\checkmark \\

Transparent and reproducible experimental framework &
Limited &
\checkmark \\

Deployment considerations (security, scalability and transparency) &
Partial &
\checkmark \\

Digital twin perspective for personalised treatment &
Rare &
\checkmark \\

Discussion of real-world clinical deployment &
Limited &
\checkmark \\

\bottomrule
\end{tabular}
\caption{Comparison of the proposed framework with representative trends in existing federated learning studies for personalised healthcare.}
\end{table}

Table 3 provides a high-level comparison between the proposed framework and representative trends in the federated learning literature for personalised healthcare. While previous studies have typically addressed privacy preservation, multimodal learning, fairness, or deployment considerations independently, relatively few have considered these aspects within a single integrated framework. The proposed approach brings together multimodal data fusion, privacy-preserving collaborative learning, fairness-aware aggregation, and comparative evaluation against a centralised baseline, while also discussing practical considerations for transparency, scalability, and future digital twin-enabled applications. Collectively, these characteristics position the framework as a step towards trustworthy and deployable federated AI for personalised breast cancer prediction, while highlighting several opportunities for future work to strengthen security mechanisms and large-scale clinical validation.

\section{Conclusion and Future work}

This study investigated the viability of FL as a privacy-preserving framework for collaborative cancer treatment modelling across multiple institutions. The results demonstrate that FL not only matched but exceeded the performance of a centralised baseline. These findings align with prior work suggesting that FL can achieve competitive model performance without requiring raw data to leave the institution, reinforcing its practical appeal for healthcare settings where data sharing is legally and ethically constrained\cite{alzubi2022cloud, myrzashova2024safeguarding}.

Although the investigation revealed that its application is far from straightforward. A number of critical considerations which can be broken down into four pillars must be addressed before it can be deployed effectively in this context: Security, Transparency, Fairness and Scalability. Though only sending weights is overall securer then sending raw data, this does not guarantee patient privacy, research has shown that shared weights can still be exploited to reconstruct or infer sensitive patient information. Therefore, strategies such as differential privacy, blockchain technology and TLS certificate encryption should also be considered for a secure pipeline. Equally important is the question of fairness. Participating institutions inevitably differ in their patient demographics, and this heterogeneity introduces the risk of a biased global model. A further complication is that, because only weights are exchanged, it becomes difficult and in some cases impossible to audit the model for bias in a transparent or measurable way. 

The use of FedGFT succeeded at solving these issues with marginally higher loss overall, its improvement in MRI prediction MAE (0.0728) and its fairer behaviour across racial subgroups suggest it is the more suitable aggregation method for this application. Qualitative evaluation of model outputs at time step four provided additional insight beyond the quantitative metrics. The FL model produced finer, more plausible tumour masks that aligned more closely with the correct region compared to those generated by the centralised model, which tended towards overly broad, less detailed segmentations. This suggests that the distributed training process may act as a form of implicit regularisation, though this hypothesis would need to be tested more rigorously. The tabular outputs performed poorly, with predictions frequently collapsing to zero. This indicates a clear architectural limitation, though it is unsurprising given the small sample size used in this study, as tabular models typically require considerably larger datasets to learn meaningful patterns and produce reliable predictions. 

The proposed digital twin roadmap, in which predictions from earlier time points are recursively fed back as inputs to forecast subsequent steps, offers a conceptually sound approach to treatment simulation. While this was demonstrated only in principle, it establishes a practical pathway toward patient-specific treatment modelling that could support clinical decision making if validated properly. From an infrastructure perspective, the limitations encountered with the Flower framework highlight an important gap between research tooling and production-ready federated systems. The merging of supernodes and client roles, the absence of end-to-end TLS between clients and supernodes, and dependency version conflicts all present non-trivial barriers to real-world deployment. 

The experiments were conducted under constrained conditions, using low-resolution scans and a reduced slice subset across only three simulated clients. While this is a practical limitation, it also means the results reported here should be considered as a lower bound on what a fully resourced deployment might achieve. Greater computational capacity, higher-resolution inputs, and a larger number of participating institutions would all be expected to improve model performance and increase the robustness of the fairness and security evaluations.

Future work should prioritise deploying the system across a realistic multi-hospital mock-up that incorporates blockchain infrastructure, with performance and security assessed at scale. Engagement with clinical stakeholders will be essential to ensure the system meets practical requirements and that model outputs are evaluated against clinically accepted standards. Incentive mechanisms to reward high-quality client contributions and penalise adversarial behaviour should also be formalised, as these will be critical to the long-term integrity of any real-world federated deployment.

% %% The Appendices part is started with the command \appendix;
% %% appendix sections are then done as normal sections
% \appendix
% \section{Example Appendix Section}


% %% For citations use: 
% %%       \cite{<label>} ==> [1]

% %%
% Example citation, See \cite{lamport94}.



% \bibliographystyle{elsarticle-harv}
% \bibliographystyle{elsarticle-num}
% \bibliography{references}


% %% For numbered reference style

%   \textit{\LaTeX: a document preparation system},
%   Addison Wesley, Massachusetts,

% \end{thebibliography}
\end{document}